\pdfoutput=1

\documentclass[11pt]{article}

\usepackage[final]{coling}

\usepackage{times}
\usepackage{latexsym}
\usepackage{multirow}
\usepackage[T1]{fontenc}
\usepackage{subfigure}
\usepackage[utf8]{inputenc}

\usepackage{microtype}

\usepackage{inconsolata}
\usepackage{tabularx, array} 
\usepackage{booktabs} 
\usepackage{amsmath}
\usepackage{graphicx}
\usepackage{pgfplots}
\pgfplotsset{compat=1.17} 
\title{Data Quality Enhancement on the Basis of Diversity with Large Language Models for Text Classification: Uncovered, Difficult, and Noisy}

\author{
 \textbf{Min Zeng},
 \textbf{Caiquan Liu},
 \textbf{Shiqi Zhang},
 \textbf{Li Xie}, 
 \textbf{Chen Sang},
  \textbf{Xiaoxin Chen}
\\
vivo AI Lab
\\
 \small{
  \href{mailto:zengmin.ai@vivo.com}{zengmin.ai@vivo.com}
 }
}

\begin{document}
\maketitle
\begin{abstract}
In recent years, the use of large language models (LLMs) for text classification has attracted widespread attention. Despite this, the classification accuracy of LLMs has not yet universally surpassed that of smaller models. LLMs can enhance their performance in text classification through fine-tuning. However, existing data quality research based on LLMs is challenging to apply directly to solve text classification problems. To further improve the performance of LLMs in classification tasks, this paper proposes a data quality enhancement (DQE) method for text classification based on LLMs. This method starts by using a greedy algorithm to select data, dividing the dataset into sampled and unsampled subsets, and then performing fine-tuning of the LLMs using the sampled data. Subsequently, this model is used to predict the outcomes for the unsampled data, categorizing incorrectly predicted data into uncovered, difficult, and noisy data. Experimental results demonstrate that our method effectively enhances the performance of LLMs in text classification tasks and significantly improves training efficiency, saving nearly half of the training time. Our method has achieved state-of-the-art performance in several open-source classification tasks.
\end{abstract}

\section{Introduction}
In the field of natural language processing (NLP), text classification is a fundamental and critical task that aims to automatically categorize given texts into predefined categories \citep{txclass}. Traditional text classification methods, such as rule-based dictionaries and machine learning algorithms, generally perform well with structured text data \citep{tx_d1, traditional}. With the advent of pre-trained language models, a significant technological breakthrough has been achieved in text classification, particularly with the introduction of the Bidirectional Encoder Representations from Transformers (BERT) \citep{bert} model. With the further development of technology, large language models (LLMs) such as GPT-4 \citep{gpt4}, LLaMA \citep{Llama,llama2}, Gemma \citep{gemma,gemma2}, GLM \citep{GLM}, and Qwen \citep{qwen,qwen2}, has demonstrated tremendous potential across various NLP tasks \citep{nlp1,nlp2}. Generative-based LLMs can perform text classification tasks by crafting specific prompts tailored to the tasks at hand. Despite their outstanding capabilities across many NLP tasks, LLMs often do not outperform smaller models in text classification \citep{e1}. Unlike BERT, the output of LLMs is often uncontrollable, which may lead to the generation of unexpected content. Through Supervised Fine-Tuning (SFT), the ability of LLMs to follow instructions can be further enhanced, thereby improving their performance in text classification tasks \citep{sfttx}.

One major challenge in fine-tuning LLMs is acquiring high-quality data. In the early days, OpenAI proposed a principle known as the Scaling Law \citep{Scalaw}, which suggested that the ultimate performance of LLMs primarily depends on the scale of three factors: computational power, model parameters, and the amount of training data. However, as AI technology continues to advance, this principle has come under scrutiny. Recent research from Microsoft indicates that the quality of training data is more critical than its quantity. The phi-1.5 \citep{phi-1.5} model outperforms Llama-7B and Llama2-7B on multiple benchmarks, despite Llama2-7B being trained on a staggering 2 trillion tokens, while phi-1.5 was trained on only 300 billion tokens. The key to phi-1.5’s success lies in the use of high-quality data. Therefore, selecting high-quality data is crucial for improving model performance.

To acquire high-quality datasets for fine-tuning LLMs for text classification, this paper introduces a data quality enhancement (DQE) approach based on LLMs. Initially, the dataset is divided into sampled and unsampled subsets using a greedy algorithm. The model is then fine-tuned on the sampled dataset to predict the results of the unsampled dataset. Data that is incorrectly predicted is further categorized into three types: uncovered data not represented in the sampled dataset, data that is difficult to fine-tune within the sampled dataset, and noisy data resulting from incorrect labeling. Subsequently, noisy data are removed from the sampled subset, and it is merged with the uncovered and difficult data from the unsampled subset to form the final dataset. The main contributions of this study are as follows:
\begin{itemize}
\item A novel approach combining LLMs with textual similarity techniques has been proposed to redefine the categorization of uncovered, difficult, and noisy data. 

\item A data quality enhancement strategy has been developed to select high-quality datasets suitable for text classification tasks. This strategy ensures data diversity and further selects uncovered and difficult data beneficial for model fine-tuning while eliminating data that impedes training, namely noisy data.

\item The method presented significantly improves the training efficiency of LLMs, achieving better fine-tuning performance with nearly half of the data volume, thus saving nearly half of the training time cost.
\end{itemize}

\section{Related Work}
Despite the powerful comprehension capabilities of LLMs due to their extensive parameterization, controlling their output can often be challenging \citep{instructionf}. Recent studies have demonstrated that SFT on datasets can effectively enhance the instruction-following abilities of LLMs \citep{instruction, Reflection-Tuning, Pushing}. Currently, some research efforts focus on using data augmentation techniques to expand the training data volume, such as generating additional training data with ChatGPT to improve the performance of text classification tasks \citep{DataGeneration, SDG}. Although these methods generally yield good results on public datasets, data and labels produced by LLMs like ChatGPT may have biases. Moreover, the quality of data is even more critical than its quantity, as high-quality data is essential for enhancing model accuracy and reliability.

\citet{IDF} proposed a guided approach to balance the quality and quantity of instructional data, allowing LLMs to identify and select the most beneficial data for training from the dataset. They introduced the Instruction-Following Difficulty (IFD) metric to locate training samples that are crucial for improving model performance. However, the IFD metric may be affected by noisy data, and high-IFD samples are not necessarily genuinely challenging samples. Moreover, their study did not consider the diversity of the data.

The complexity, quality, and diversity of data are widely regarded as crucial factors in enhancing model performance. \citet{D4} from Meta AI Research proposed the Document De-Duplication and Diversification (D4) method, which balances data diversity and quantity using the K-Means algorithm and adjusting data selection proportions. \citet{mods} introduced a model-oriented data selection (MODS) method, which considers three aspects: data quality, coverage, and necessity. This method uses a reward model to evaluate data quality, applies the K-center-greedy algorithm to maximize data diversity, and includes data that the model has not yet learned, offering a comprehensive approach. However, this method depends on a fragile reward value model, leading to potential biases in selected data. To address this issue, \citet{CAR} introduced a scoring model aligned with expert preferences to precisely evaluate data quality and use clustering models like k-Means to maintain diversity among candidate data. Although these studies thoroughly consider data diversity, they fail to further identify the uncovered and difficult data required for model training, nor do they effectively remove noisy data that impedes model training. In text classification tasks, noisy data due to labeling errors is particularly challenging to avoid.

\begin{figure*}[htbp]
\centerline{\includegraphics[width=0.95\textwidth]{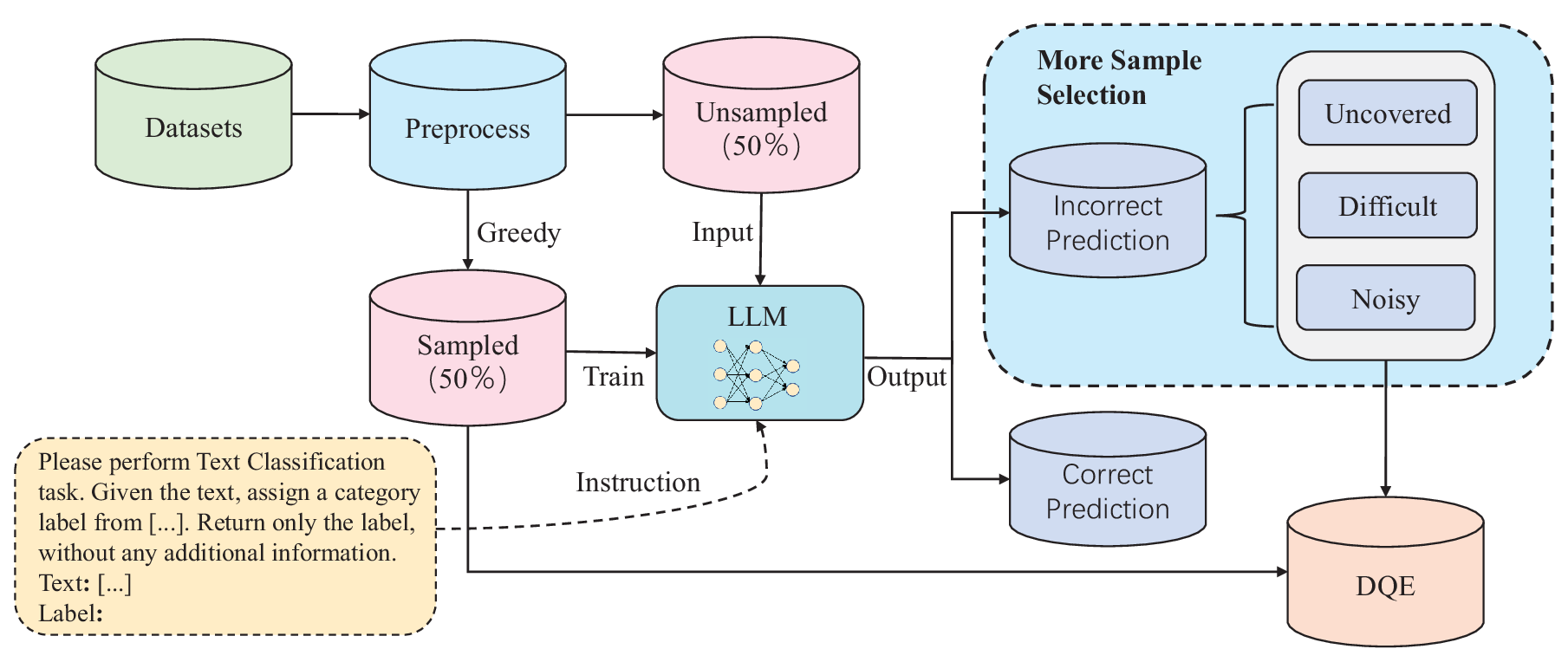}}
\caption{The overall structure of the DQE method.}
\label{HQDS}
\end{figure*}

\section{Methodology}
Existing work on data quality based on LLMs primarily focuses on pre-training data and fine-tuning data for generic model commands, but these methods are not directly applicable to text classification tasks based on LLMs. This paper proposes a new method that combines cosine similarity to identify uncovered, difficult, and noisy samples, thereby improving the training and prediction effectiveness of LLMs. Initially, the dataset undergoes preliminary processing, including the removal of duplicate data, handling of missing values, and cleaning of inconsistent labels. After preprocessing, as shown in Figure~\ref{HQDS}, we convert the data into vector representations using a vector model and sample using the K-Center-Greedy \citep{kgreedy} algorithm. It is important to note that too little sampling may lead to a significant amount of uncovered data, whereas too much sampling might increase the likelihood of selecting noisy data. Therefore, we adopt a more balanced sampling approach, collecting half of the data as sampled and leaving the remainder as unsampled. Subsequently, we fine-tune the LLM using the sampled data to predict the unsampled data. During the prediction of unsampled data, incorrectly predicted samples can be further categorized into uncovered, difficult, and noisy samples.

\subsection{Preprocessing}
Before sampling the data, the dataset undergoes a simple preprocessing step. First, we check and remove the missing values in the dataset, where missing values refer to abnormal data with only text and no labels or no text and only labels, and the presence of missing values will interfere with the training results of the model. Next, we remove duplicate data, as duplicates have limited significance for model training and can impact subsequent data sampling. Finally, we identify samples with inconsistent labels in the dataset, as inconsistent labels can cause difficulties in model convergence during training.

\subsection{Greedy Sampling}
After preprocessing the dataset, we first employ the all-mpnet-base-v2\footnote{\url{https://huggingface.co/sentence-transformers/all-mpnet-base-v2}} model to convert text content from the dataset into vectors. This model is capable of transforming text into vectors rich with semantic information, which facilitates the computation of semantic distances between texts. Subsequently, we utilize the K-Center-Greedy algorithm to select half of the data from the original dataset. It starts by randomly selecting a sample as the initial vector center and then iteratively selects the sample that is farthest from the current vector center to add to the dataset and update the center. This process continues until the predetermined number of K samples is selected. To maintain a balance between the sampled and unsampled subsets, K is generally set to half of the size of the pre-processed training dataset. The data sampled via the K-Center-Greedy algorithm typically covers a broader range, effectively ensuring the diversity and representativeness of the sampled data.

\begin{figure*}[htbp]
\centerline{\includegraphics[width=0.9\textwidth]{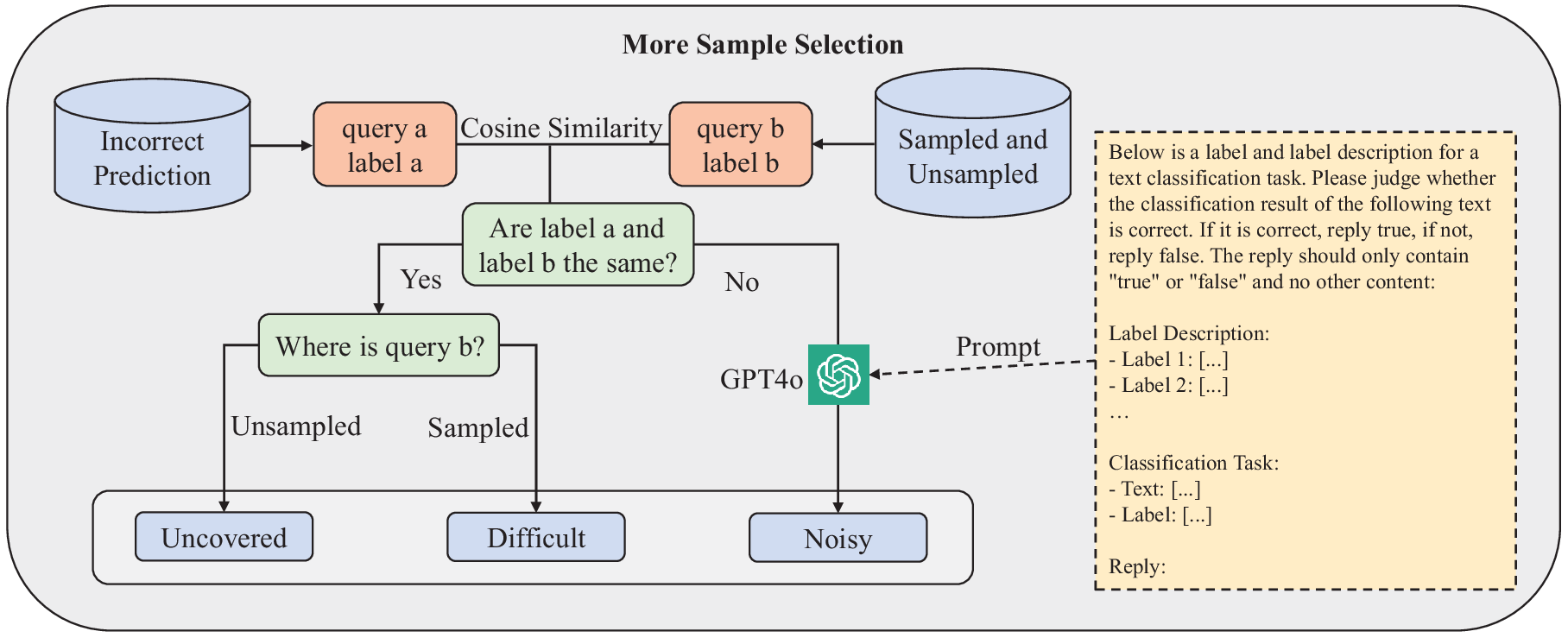}}
\caption{The identification process of Uncovered, Difficulty, and Noisy.}
\label{enhancement}
\end{figure*}

\subsection{More Sample Selection}
This study aims to further select data that is beneficial for model training on the basis of data diversity while eliminating data that hinders model training. We facilitate the training process by incorporating uncovered data and increasing difficult samples. However, noisy data can disrupt the model's convergence process, thereby impacting performance. Especially in text classification tasks, training a deep learning model usually requires a large amount of annotated data, and label noise during the annotation process is a common and unavoidable issue.

To further enhance data quality, after the second stage of greedy sampling, we divide the dataset into sampled and unsampled subsets. We employ a pre-trained LLM to fine-tune the sampled dataset under supervision and then predict the unsampled dataset. As shown in Figure~\ref{enhancement}, for each incorrectly predicted data in the unsampled dataset, we use cosine similarity to find the most similar sample in the entire dataset. The calculation process of the maximum similarity can be expressed as follows:
\begin{equation}
q_b = \underset{q_k \in Q}{\arg\max}\cos (q_a, q_k)
\end{equation}
where $q_a$ represents query a, and $Q$ represents the data in the entire training set consisting of queries except $q_a$. $q_b$ represents the query with the maximum cosine similarity in $Q$. Typically, two highly similar data items have the same label. Based on this method, we categorize the incorrectly predicted data into the following three types:

1) \textbf{Uncovered}: If the incorrectly predicted sample has the same label as the most similar sample found and this most similar data is in the unsampled dataset, it is considered that the features related to this data might not have been involved in the model's fine-tuning within the sampled dataset, i.e., it was not covered by the sampled dataset. Although the sampled dataset is obtained through greedy sampling of semantic vectors, it is a normal phenomenon that the unsampled dataset contains data not covered by the sampled dataset when the original dataset has high diversity. In this case, we will add this data to the final training set.

2) \textbf{Difficult}: If the incorrectly predicted sample has the same label as the most similar sample found, and this most similar sample is in the sampled dataset, this indicates that although the related features have been trained, the model failed to predict the outcome correctly, defining these samples as difficult. Difficult samples are often challenging to learn due to complex features or insufficient numbers of related category samples. For these cases, we also add these data to the final training set.

3) \textbf{Noisy}: If the incorrectly predicted sample has different labels compared to the most similar sample found, i.e., two similar data are labeled with two different labels, then these two data are suspected to be noisy data. Noisy data is a common problem in text classification tasks. The main sources of noisy data are that most datasets for text classification tasks are obtained through manual or model annotations, and whether it is manual or model annotation, both can be influenced by subjectivity. In the face of a large dataset, it is difficult to avoid labeling errors. Another factor is that some data descriptions are vague, making it hard to determine which label the content belongs to. These data can cause the model to have difficulty converging during training. To address this issue, we include descriptions for each category in the prompts and use the GPT-4o model to assist in judging the labels. Once GPT-4o determines that a label is incorrect, we will remove that data.

\section{Experimental Setup}

\subsection{Datasets}
To ensure the openness and transparency of our experimental results, we have selected a series of widely used public text classification datasets for this study. Additionally, to establish a competitive benchmark, we have chosen the state-of-the-art results from the PapersWithCode\footnote{\url{https://paperswithcode.com/sota}} as the baseline for this experiment, most of which are derived from traditional smaller models. Below is a detailed introduction to the text classification datasets used in this experiment:

\begin{itemize}
\item \textbf{MR} \citep{MR} is a binary classification task about movie reviews, the data can be divided into positive and negative according to sentiment, containing 8530 training samples and 1066 test samples.

\item  \textbf{CR} \citep{CR} is a binary classification task about customer reviews, the data can be divided into positive and negative according to sentiment, containing 3394 training samples and 376 test samples. 

\item  \textbf{IMDb} \citep{imdb} is a binary classification task about movie reviews, the data can be divided into positive and negative according to sentiment, containing 25000 training samples and 25000 test samples. 

\item \textbf{SST-2} \citep{SST} is a binary classification task about movie reviews, the data can be divided into positive and negative according to sentiment, containing 67349 training samples and 1821 test samples. 

\item \textbf{SST-5} \citep{SST} is a five-class classification task about movie reviews, the data can be divided into very negative, negative, neutral, positive, and very positive, containing 8544 training samples and 2210 test samples.

\item  \textbf{AG News} \citep{AGNEWS} is a four-class classification task about news, which can be divided into World, Sports, Business, and Sci/Tech, containing 120000 training samples and 7600 test samples.
\end{itemize}

\subsection{Large Language Models}

The experiments mainly involve the following two publicly available LLMs:

\vspace{10pt}
\noindent \textbf{GPT-4o\footnote{\url{https://platform.openai.com/docs/models/gpt-4o}}} is an optimized version of GPT-4, designed specifically to improve efficiency and performance. It inherits the core architecture of GPT-4 with optimizations, particularly showing significant improvements in inference speed and resource utilization. GPT-4o is capable of handling various complex natural language processing tasks. It is primarily used in experiments for detecting noisy data and can be accessed through the official API provided by OpenAI.

\vspace{10pt}
\noindent \textbf{Qwen2-7B-Instruct\footnote{\url{https://huggingface.co/Qwen/Qwen2-7B-Instruct}}} is Alibaba's latest open-source LLM, capable of understanding multiple languages. It has 7 billion parameters and supports a context length of 131,072 tokens, performing exceptionally well across various evaluation benchmarks. In this experiment, Qwen2-7B-Instruct is chosen as the base model for fine-tuning.

\subsection{Implementation Details}
We deployed a multi-node multi-GPU server environment, conducting experiments on two servers each equipped with four L40s GPUs. Each L40s GPU has 48GB of VRAM. We employed the second stage of DeepSpeed's \citep{zero} data parallel training strategy, distributing the data evenly across different servers to facilitate distributed training and accelerate the training process. During the fine-tuning phase, we performed full-parameter fine-tuning using the open-source framework Swift, setting the learning rate to 1e-5 and the batch size to 1. To increase the effective batch size, we utilized gradient accumulation techniques with the number of accumulation steps set to 16. Depending on the dataset size, we dynamically adjusted the training epochs for each experiment, ranging from 1 to 3.

\setlength{\tabcolsep}{3.8pt}
\begin{table}
  \centering
  \begin{tabular}{ccccc}
    \hline
    \textbf{Dataset}& \textbf{Test}&\textbf{Full-Data}&\textbf{Greedy} &\textbf{DQE} \\
    \hline
    MR & 1066  & 8530& 4265 &  \textbf{4351}  \\
    CR & 376 & 3394& 1694 & \textbf{1720}  \\
    IMDb  & 25000 & 25000 & 12452 & \textbf{12574}  \\    
    SST-2  & 1821  & 67349 & 33486&   \textbf{33426}  \\
    SST-5  & 2210    & 8544  & 4267 &  \textbf{4514} \\
    AG News  & 7600  & 120000& 60000 & \textbf{60375}   \\    
    \hline
  \end{tabular}
  \caption{\label{sample}
    For the sampling results of each dataset: The number of test sets and training sets corresponding to each data set, and the amount of data under greedy sampling and DQE sampling.
  }
\end{table}

\begin{table*}
  \centering
  \begin{tabular}{cccccc}
    \hline
    \textbf{Dataset}&\textbf{Baseline} &\textbf{Base-Model} &\textbf{Full-Data}&\textbf{Greedy}  &\textbf{DQE} \\
    \hline
    MR    & 93.30  & 76.83 &  92.68& 92.96& \textbf{93.81} \\  
    CR   &  93.54 & 80.85 & 94.95 &94.95 & \textbf{95.48} \\
    IMDb    &  96.68 & 90.26 &  97.32 & 97.38& \textbf{97.86} \\
    SST-2    & 97.50  & 83.69 &  97.47&97.75& \textbf{98.35} \\
    SST-5    & 59.80  & 43.80 &  61.62&60.90& \textbf{61.95} \\
    AG News    & 85.00  & 69.64 & 95.38 &95.04&\textbf{95.70} \\
    \hline
  \end{tabular}
  \caption{\label{result}
   Accuracy (\%) of different methods on the test set.
  }
\end{table*}

\begin{table}
  \centering
  \begin{tabular}{ccc}
    \hline
     \textbf{Dataset} & \textbf{T-Statistic} &\textbf{P-Value} \\
    \hline
    MR     & 3.48 & 0.0005 \\
    CR    & 1.42 & 0.1576 \\
    IMDb    &11.65 & < 0.0001\small{(2.76e-13)}\\
    SST-2    &4.02 & < 0.0001\small{(6.14e-05)} \\
    SST-5   & 2.65 & 0.0081\\
    AG News  & 4.91 & < 0.0001\small{(9.47e-7)}\\
    \hline
  \end{tabular}
  \caption{
    Significance test results of DQE and Full-Data}
  \label{Significance}
\end{table}

\subsection{Results}
In addition to comparing with the baseline results, we also compared the DQE method against methods using full data and greedy sampling. Table~\ref{sample} shows the sizes of the datasets used during the experiment. Full-Data represents using the entire dataset as the training set, while Greedy refers to the subset obtained after preprocessing the full dataset and collecting half of it using the K-Center-greedy algorithm. DQE is the final sampling result obtained after further processing uncovered, difficult, and noisy data on the basis of the Greedy. From the table, it can be observed that the final sampling result of DQE is about 50\% of the Full-Data, with the sampling outcome for the SST-2 dataset even falling below 50\%.

As shown in Table~\ref{result}, we measured model performance using accuracy. To ensure the fairness of the experiment, the models fine-tuned for the Full-Data, Greedy, and DQE datasets are all based on the same hyperparameters. The DQE method proposed in this article achieved the best results in all tasks. The baseline results are mainly based on traditional smaller models, and their performances along with other related research can be found in the PapersWithCode. The Base-Model in this paper refers to the direct operation results of the unoptimized Qwen2-7B-Instruct model. Except for IMDb, the un-tuned base model performed poorly in other tasks, making it the weakest overall, not surpassing the baseline in any of the six tasks. In contrast, models fine-tuned with full data surpassed the baseline in four tasks: CR, IMDb, SST-5, and AG News, but not in MR and SST-2. Models fine-tuned with greedy sampled data exceeded the baseline in CR, IMDb, SST-2, and AG News, but not in MR and SST-5.


In our research, we employed the t-test to evaluate the statistical significance of performance differences between models. Specifically, we assigned a value of 1 to correctly predicted samples and a value of 0 to incorrectly predicted ones, using this scheme to quantitatively assess the models' performance across various datasets. For DQE and the Full-Data, we further explored whether there is a significant performance difference between the two on different data sets. Our findings revealed that although the accuracy of models obtained through DQE and Full-Data was comparable on the test sets, their differences were statistically significant. Table~\ref{Significance} presents the results of significance tests across varying datasets. For the MR, IMDb, SST-2, and AG News datasets (with respective sample sizes of 1066, 25000, 1821, and 7600), the t-statistics were 3.482, 11.650, 4.017, and 4.906, paired with p-values significantly below the threshold of 0.05, indicating that the performance improvements demonstrated by DQE were statistically significant. However, for the CR dataset with only 376 samples, despite a t-statistic of 1.416, the corresponding p-value of 0.1575 indicated that the performance difference between the two methods did not reach a level of significance in smaller-scale datasets, possibly due to the reduced power of statistical tests. For the SST-5 dataset with a sample size of 2210, the t-statistic was 2.649, and the p-value was 0.0081, also reflecting a significant performance disparity between the two methods. Therefore, we conclude that while DQE and Full-Data methods may exhibit similar levels of accuracy, the DQE approach proposed in this study demonstrates a significant performance enhancement over Full-Data across the majority of the datasets tested.

\begin{figure}[htbp]
 \hspace{0.15cm}
\centerline{\includegraphics[width=0.51\textwidth]{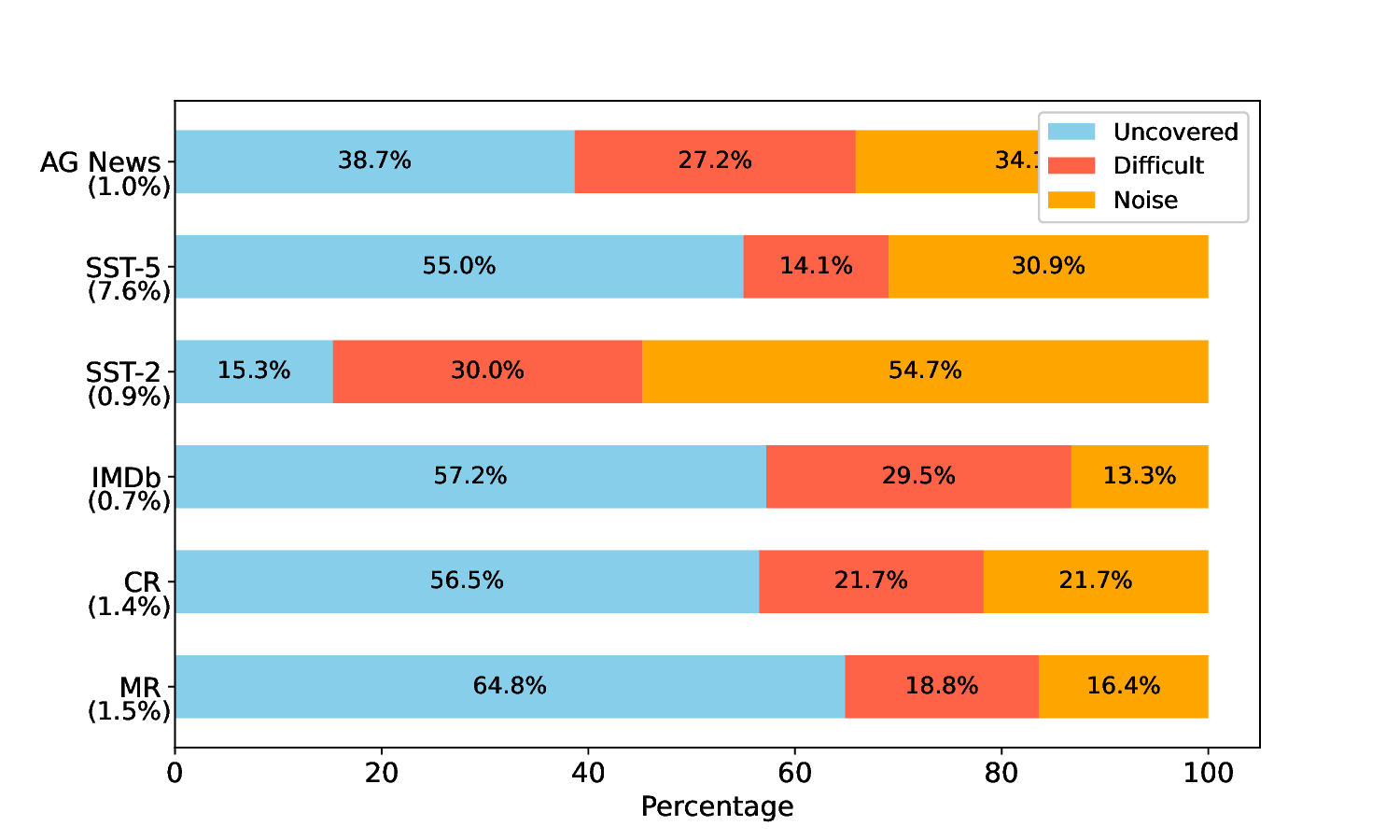}}
\caption{the proportion between uncovered, difficult, and noisy data, as well as their combined proportion in the entire training set.}
\label{count}
\end{figure}

The development of LLMs generally adheres to the Scaling Law. However, this principle is not always applicable, especially in classification tasks. In fact, increasing data size is not always beneficial, as large datasets may give rise to issues such as data redundancy and noise caused by inconsistent labeling. Comparing the results of full data and greedy sampling, greedy sampling showed higher accuracy in the test sets of MR, IMDb, and SST-2 than full data, and matched performance on CR. This is because greedy sampling ensures a broad coverage by selecting data with the greatest semantic distances, and the probability of label inconsistencies is lower due to the significant semantic distances, thus offering better predictive performance. In the presence of numerous similar noisy data, the results of training on the entire dataset can be easily disrupted by the noisy data. In contrast, the greedy sampling method would significantly reduce the amount of noisy data selected in such scenarios. However, greedy sampling did not outperform in SST-5 and AG News, possibly because it only collected half the samples, potentially leaving some data uncovered, including difficult samples, and may be mixed with noisy data that exist alone.

\begin{figure*}[htbp]
\centerline{\includegraphics[width=0.95\textwidth]{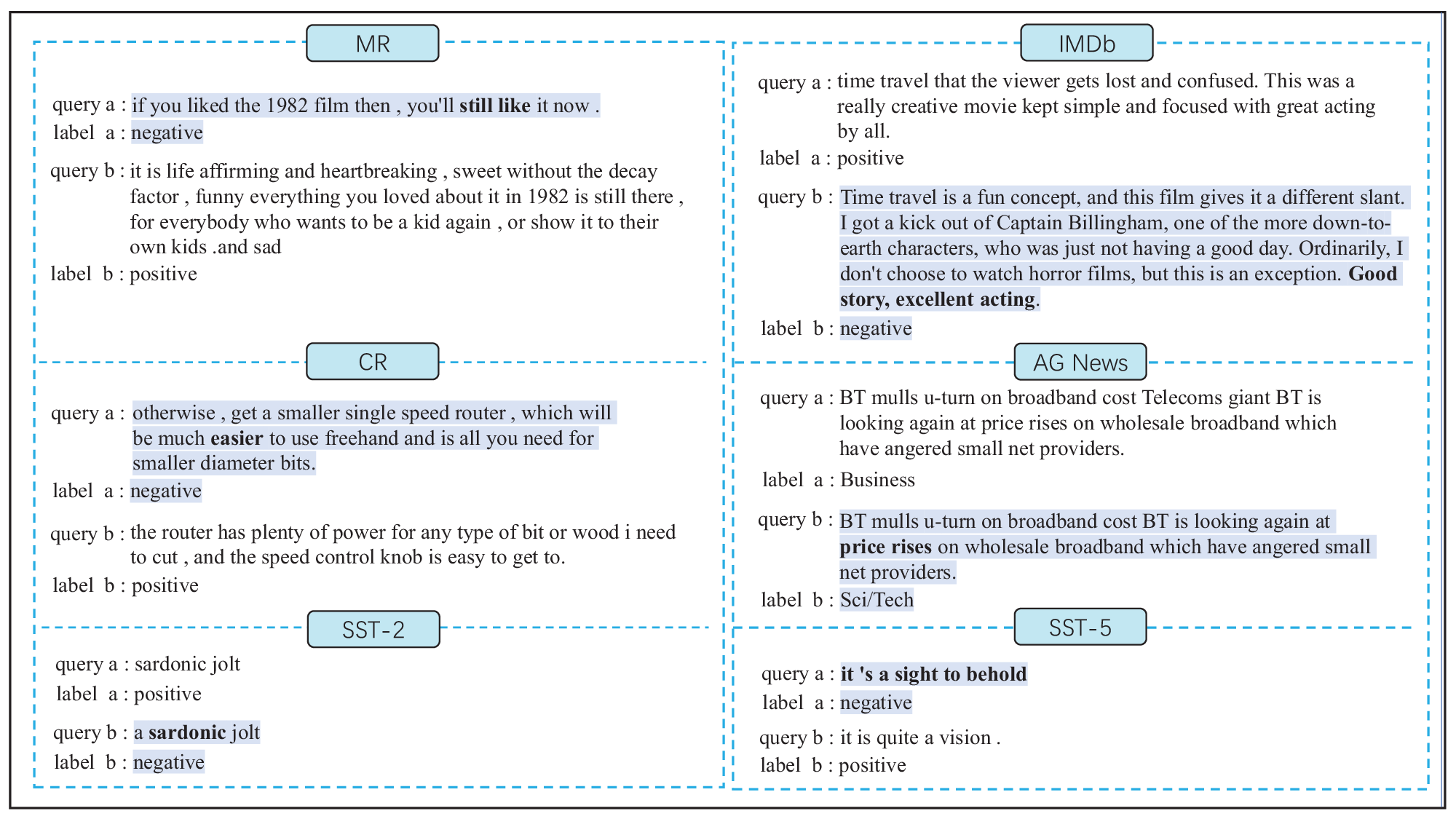}}
\caption{Example of noisy data found by the DQE.}
\label{trainnoise}
\end{figure*}

The DQE method further identifies uncovered, difficult, and noisy data based on greedy sampling, thereby expanding the coverage of sampling results and enhancing learning from difficult samples, while also removing some suspected noisy data. Figure~\ref{count} shows the proportions of uncovered, difficult, and noisy data identified during the sampling process by the DQE method, as well as their collective proportion in the entire training set. Notably, in the IMDb dataset, these identified data types account for the smallest proportion, and DQE improved the accuracy rate in the test set by 0.48\% compared to Greedy. Conversely, in SST-5, these data types constitute the largest proportion, 7.6\% of the total data, and correspondingly, the accuracy improvement of DQE compared to Greedy in the test set is the most significant, exceeding 1\%. These results indicate that the method proposed in this paper effectively enhances model performance based on greedy sampling. Compared to using the full dataset, the data selected by DQE not only covers a broader range but also effectively mitigates noise issues, enabling DQE to surpass the full dataset's accuracy in the test set using only about half of the data, thereby significantly enhancing training efficiency.

\begin{table}
  \centering
  \begin{tabular}{clcc}
    \hline
     \textbf{Dataset} & \textbf{Base-Model} & \textbf{Full-Data} &\textbf{DQE} \\
    \hline
    MR   & 156 \small(14.63\%)   & 0  & 0\\
    CR   & 56 \small(14.89\%)  & 0  & 0 \\
    IMDb   & 1094 \small(4.2\%)   & 0  & 0\\
    SST-2   & 227 \small(12.47\%) & 0  & 0\\
    SST-5   & 2 \small(0.09\%)  & 0  & 0\\
    AG News   & 167 \small(2.20\%) & 0  & 0\\
    \hline
  \end{tabular}
  \caption{
    The amount of data beyond expectations in the model output corresponds to each method.}
  \label{instruction}
\end{table}

\section{Analysis and Discussion}

\subsection{Instruction Following Capability}
A major challenge of generative LLMs for classification tasks is ensuring that the outputs content as expected. Unlike BERT models, generative models might produce outputs that are outside of expectations. Table~\ref{instruction} shows the number of instances where the output of the LLM did not meet expectations and their proportion in the entire test set. The experimental results indicate that the un-tuned LLM sometimes produces unexpected outcomes, such as outputs that do not adhere to the specified format or are irrelevant labels. Particularly in the MR, CR, and SST-2 datasets, the proportion of data with output issues exceeds 10\%. These datasets are binary classification problems, but binary classification does not seem to effectively segment these data, leading to erroneous output categories. In contrast, in the SST-5 dataset, the proportion of problematic data is only 0.5\%, and the five categories effectively segment the dataset, with issues mainly due to non-standard output formats. Models fine-tuned with Full-Data and DQE significantly enhance the command-following capabilities of the LLM, ensuring that outputs meet expectations.

\subsection{Noisy Data}

Experimental results indicate that even widely used public datasets commonly contain noisy data, especially within larger datasets where the issue of noisy data is more pronounced. Figure~\ref{trainnoise} displays examples of data suspected to have incorrect labels identified during our experiments. "query a" refers to data from the unsampled dataset where the predictions by the model, after fine-tuning with the sampled dataset, do not align with the actual labels; "query b" refers to data identified through cosine similarity as having the highest similarity to "query a". "label a" and "label b" represent their respective labels. In the SST-2 and AG News datasets, we found a significant number of such highly similar noisy data. It is evident that there is a high probability of label errors in these pairs of data. For such data, we conducted further verification using GPT-4o, if GPT-4o is unable to determine or determines that the labeling is incorrect, then that data is discarded.

\begin{figure*}[htbp]
\centerline{\includegraphics[width=0.8\textwidth]{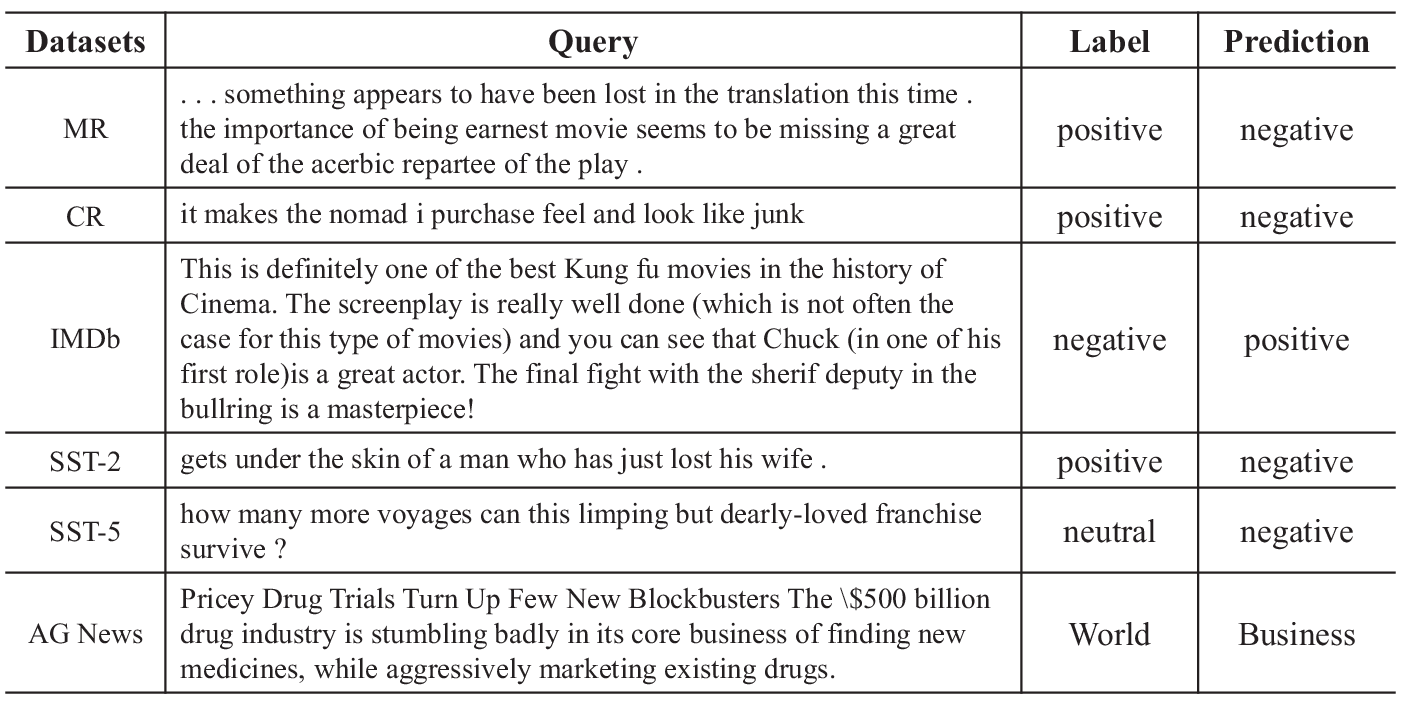}}
\caption{Examples of prediction results of DQE on the test set.}
\label{analysis result}
\end{figure*}

In addition to the noisy data analysis in the training set, we also conducted further analysis of the performance of DQE in the test set. By analyzing a randomly selected portion of the data with incorrect model predictions from each dataset, we discovered that some of these errors were not due to inaccuracies in model predictions but were caused by incorrect labels on the data itself. Figure~\ref{analysis result} presents examples of DQE's prediction results across various test sets, revealing that label inaccuracies occur in the test sets of each dataset. This indicates that even carefully selected test sets cannot completely avoid noise issues. Therefore, in practical applications, the predictive performance of models sampled using the DQE method is expected to be higher than what is displayed in the test sets.

\subsection{Discussion}
It's worth noting that in the data sampling process, this study divides the dataset into sampled and unsampled portions using a greedy algorithm. Further, we categorize the unsampled subset into uncovered, difficult, and noisy classes based on text similarity. Next, let's analyze the sampled dataset: since this data will be used in the final training set, it does not contain Uncovered instances. Regarding difficulty, poorly performing samples from the unsampled subset are transferred to the sampled subset, thereby partially mitigating difficulty issues within the sampled data. As for noisy data, most paired noisy instances can be identified between the sampled and unsampled subsets using DQE. Due to the use of a greedy sampling strategy, the probability of encountering paired similar noisy data within the sampled subset is relatively low. 

Additionally, this paper identifies uncovered, difficult, and noisy data based on the top 1 recall without setting a specific similarity threshold.  This approach is adopted because our sampled data is obtained through greedy sampling of semantic distances. This means that if a particular data item has a significantly larger semantic distance compared to the overall dataset, it would have a higher probability of being included in the sampled. Therefore, in the unsampled subset, most data items will have a sufficiently similar counterpart. Nevertheless, in some special cases, a similarity threshold can be set depending on the specific situation of the embedding model.

\section{Conclusion}
In this paper, we introduce DQE, a data quality enhancement method based on LLM for text classification. This method ensures data diversity while further identifying uncovered and difficult data that can improve model performance, and simultaneously identifies noisy data detrimental to model training. Experimental results show that compared to traditional greedy sampling, the DQE method effectively identifies uncovered, difficult, and noisy data to enhance performance. By using only about half of the data, the DQE method surpasses the performance of the full dataset, significantly improving accuracy in classification tasks and saving half of the training time, thereby greatly enhancing training efficiency. Moreover, in comparative experiments with the baseline model, the DQE method achieved state-of-the-art results.

\section{Limitations}
The following limitations exist in this study:
\begin{itemize}
\item The vector model used in the experiments has good general capabilities and therefore was not fine-tuned. In practical applications, fine-tuning the vector model for specific text classification tasks could yield more accurate semantic vectors and better suit specific requirements.
\item This work focuses on improving the quality of training data and has demonstrated superior performance on public datasets accordingly. As such, we did not optimize the prompt words used in the experiments, nor did we conduct exhaustive hyperparameter tuning, both of which could further enhance performance in text classification tasks.
\item The method introduced in this paper enhances data quality while aiming to detect noisy data that has been incorrectly labeled as extensively as possible. However, it is not capable of identifying all mislabeled data. Currently, there is no known method that can guarantee the detection of all noisy data in every dataset.
\end{itemize}

\bibliography{custom} 

\begin{thebibliography}{35}
\providecommand{\natexlab}[1]{#1}

\bibitem[{Achiam et~al.(2023)Achiam, Adler, Agarwal, Ahmad, Akkaya, Aleman, Almeida, Altenschmidt, Altman, Anadkat et~al.}]{gpt4}
Josh Achiam, Steven Adler, Sandhini Agarwal, Lama Ahmad, Ilge Akkaya, Florencia~Leoni Aleman, Diogo Almeida, Janko Altenschmidt, Sam Altman, Shyamal Anadkat, et~al. 2023.
\newblock \href {https://arxiv.org/pdf/2303.08774} {Gpt-4 technical report}.
\newblock \emph{arXiv preprint arXiv:2303.08774}.

\bibitem[{Bai et~al.(2023)Bai, Bai, Chu, Cui, Dang, Deng, Fan, Ge, Han, Huang et~al.}]{qwen}
Jinze Bai, Shuai Bai, Yunfei Chu, Zeyu Cui, Kai Dang, Xiaodong Deng, Yang Fan, Wenbin Ge, Yu~Han, Fei Huang, et~al. 2023.
\newblock \href {https://arxiv.org/pdf/2309.16609} {Qwen technical report}.
\newblock \emph{arXiv preprint arXiv:2309.16609}.

\bibitem[{Bucher and Martini(2024)}]{sfttx}
Martin Juan~Jos{\'e} Bucher and Marco Martini. 2024.
\newblock \href {https://arxiv.org/pdf/2406.08660} {Fine-tuned'small'llms (still) significantly outperform zero-shot generative ai models in text classification}.
\newblock \emph{arXiv preprint arXiv:2406.08660}.

\bibitem[{Devlin(2018)}]{bert}
Jacob Devlin. 2018.
\newblock \href {https://aclanthology.org/N19-1423.pdf} {Bert: Pre-training of deep bidirectional transformers for language understanding}.
\newblock \emph{arXiv preprint arXiv:1810.04805}.

\bibitem[{Du et~al.(2023)Du, Zong, and Zhang}]{mods}
Qianlong Du, Chengqing Zong, and Jiajun Zhang. 2023.
\newblock Mods: Model-oriented data selection for instruction tuning.
\newblock \emph{arXiv preprint arXiv:2311.15653}.

\bibitem[{Du et~al.(2021)Du, Qian, Liu, Ding, Qiu, Yang, and Tang}]{GLM}
Zhengxiao Du, Yujie Qian, Xiao Liu, Ming Ding, Jiezhong Qiu, Zhilin Yang, and Jie Tang. 2021.
\newblock \href {https://arxiv.org/pdf/2103.10360} {Glm: General language model pretraining with autoregressive blank infilling}.
\newblock \emph{arXiv preprint arXiv:2103.10360}.

\bibitem[{Edwards and Camacho-Collados(2024)}]{e1}
Aleksandra Edwards and Jose Camacho-Collados. 2024.
\newblock \href {https://arxiv.org/pdf/2403.17661} {Language models for text classification: Is in-context learning enough?}
\newblock \emph{arXiv preprint arXiv:2403.17661}.

\bibitem[{Ge et~al.(2024)Ge, Liu, Hu, Meng, Tao, Zhao, Ma, Zhang, Yang, and Xiao}]{CAR}
Yuan Ge, Yilun Liu, Chi Hu, Weibin Meng, Shimin Tao, Xiaofeng Zhao, Hongxia Ma, Li~Zhang, Hao Yang, and Tong Xiao. 2024.
\newblock \href {Clustering and Ranking: Diversity-preserved Instruction Selection through Expert-aligned Quality Estimation} {Clustering and ranking: Diversity-preserved instruction selection through expert-aligned quality estimation}.
\newblock \emph{arXiv preprint arXiv:2402.18191}.

\bibitem[{Hu and Liu(2004)}]{CR}
Minqing Hu and Bing Liu. 2004.
\newblock \href {https://citeseerx.ist.psu.edu/document?repid=rep1&type=pdf&doi=299a26b7c84b532878eb4b4ff45003042f6f9423} {Mining and summarizing customer reviews}.
\newblock In \emph{Proceedings of the tenth ACM SIGKDD international conference on Knowledge discovery and data mining}, pages 168--177.

\bibitem[{Kaplan et~al.(2020)Kaplan, McCandlish, Henighan, Brown, Chess, Child, Gray, Radford, Wu, and Amodei}]{Scalaw}
Jared Kaplan, Sam McCandlish, Tom Henighan, Tom~B Brown, Benjamin Chess, Rewon Child, Scott Gray, Alec Radford, Jeffrey Wu, and Dario Amodei. 2020.
\newblock \href {https://arxiv.org/pdf/2001.08361} {Scaling laws for neural language models}.
\newblock \emph{arXiv preprint arXiv:2001.08361}.

\bibitem[{Kowsari et~al.(2019)Kowsari, Jafari~Meimandi, Heidarysafa, Mendu, Barnes, and Brown}]{tx_d1}
Kamran Kowsari, Kiana Jafari~Meimandi, Mojtaba Heidarysafa, Sanjana Mendu, Laura Barnes, and Donald Brown. 2019.
\newblock \href {https://www.mdpi.com/2078-2489/10/4/150?source=post_page---------------------------} {Text classification algorithms: A survey}.
\newblock \emph{Information}, 10(4):150.

\bibitem[{Li et~al.(2023{\natexlab{a}})Li, Chen, Chen, He, and Zhou}]{Reflection-Tuning}
Ming Li, Lichang Chen, Jiuhai Chen, Shwai He, and Tianyi Zhou. 2023{\natexlab{a}}.
\newblock \href {https://openreview.net/forum?id=xaqoZZqkPU} {Reflection-tuning: Recycling data for better instruction-tuning}.
\newblock In \emph{NeurIPS 2023 Workshop on Instruction Tuning and Instruction Following}.

\bibitem[{Li et~al.(2024{\natexlab{a}})Li, Zhang, Li, Chen, Chen, Cheng, Wang, Zhou, and Xiao}]{IDF}
Ming Li, Yong Zhang, Zhitao Li, Jiuhai Chen, Lichang Chen, Ning Cheng, Jianzong Wang, Tianyi Zhou, and Jing Xiao. 2024{\natexlab{a}}.
\newblock \href {https://aclanthology.org/2024.naacl-long.421} {From quantity to quality: Boosting {LLM} performance with self-guided data selection for instruction tuning}.
\newblock In \emph{Proceedings of the 2024 Conference of the North American Chapter of the Association for Computational Linguistics: Human Language Technologies (Volume 1: Long Papers)}, pages 7595--7628, Mexico City, Mexico. Association for Computational Linguistics.

\bibitem[{Li et~al.(2022)Li, Peng, Li, Xia, Yang, Sun, Yu, and He}]{traditional}
Qian Li, Hao Peng, Jianxin Li, Congying Xia, Renyu Yang, Lichao Sun, Philip~S Yu, and Lifang He. 2022.
\newblock A survey on text classification: From traditional to deep learning.
\newblock \emph{ACM Transactions on Intelligent Systems and Technology (TIST)}, 13(2):1--41.

\bibitem[{Li et~al.(2024{\natexlab{b}})Li, Bonatti, Abdali, Wagle, and Koishida}]{DataGeneration}
Yinheng Li, Rogerio Bonatti, Sara Abdali, Justin Wagle, and Kazuhito Koishida. 2024{\natexlab{b}}.
\newblock \href {https://arxiv.org/pdf/2407.12813} {Data generation using large language models for text classification: An empirical case study}.
\newblock \emph{arXiv preprint arXiv:2407.12813}.

\bibitem[{Li et~al.(2023{\natexlab{b}})Li, Bubeck, Eldan, Del~Giorno, Gunasekar, and Lee}]{phi-1.5}
Yuanzhi Li, S{\'e}bastien Bubeck, Ronen Eldan, Allie Del~Giorno, Suriya Gunasekar, and Yin~Tat Lee. 2023{\natexlab{b}}.
\newblock \href {https://arxiv.org/pdf/2309.05463} {Textbooks are all you need ii: phi-1.5 technical report}.
\newblock \emph{arXiv preprint arXiv:2309.05463}.

\bibitem[{Li et~al.(2023{\natexlab{c}})Li, Zhu, Lu, and Yin}]{SDG}
Zhuoyan Li, Hangxiao Zhu, Zhuoran Lu, and Ming Yin. 2023{\natexlab{c}}.
\newblock \href {https://arxiv.org/pdf/2310.07849} {Synthetic data generation with large language models for text classification: Potential and limitations}.
\newblock \emph{arXiv preprint arXiv:2310.07849}.

\bibitem[{Maas et~al.(2011)Maas, Daly, Pham, Huang, Ng, and Potts}]{imdb}
Andrew Maas, Raymond~E Daly, Peter~T Pham, Dan Huang, Andrew~Y Ng, and Christopher Potts. 2011.
\newblock \href {https://aclanthology.org/P11-1015/} {Learning word vectors for sentiment analysis}.
\newblock In \emph{Proceedings of the 49th annual meeting of the association for computational linguistics: Human language technologies}, pages 142--150.

\bibitem[{Minaee et~al.(2021)Minaee, Kalchbrenner, Cambria, Nikzad, Chenaghlu, and Gao}]{txclass}
Shervin Minaee, Nal Kalchbrenner, Erik Cambria, Narjes Nikzad, Meysam Chenaghlu, and Jianfeng Gao. 2021.
\newblock Deep learning--based text classification: a comprehensive review.
\newblock \emph{ACM computing surveys (CSUR)}, 54(3):1--40.

\bibitem[{Minaee et~al.(2024)Minaee, Mikolov, Nikzad, Chenaghlu, Socher, Amatriain, and Gao}]{nlp2}
Shervin Minaee, Tomas Mikolov, Narjes Nikzad, Meysam Chenaghlu, Richard Socher, Xavier Amatriain, and Jianfeng Gao. 2024.
\newblock \href {https://arxiv.org/pdf/2402.06196} {Large language models: A survey}.
\newblock \emph{arXiv preprint arXiv:2402.06196}.

\bibitem[{Pang and Lee(2005)}]{MR}
Bo~Pang and Lillian Lee. 2005.
\newblock \href {https://arxiv.org/pdf/cs/0506075} {Seeing stars: Exploiting class relationships for sentiment categorization with respect to rating scales}.
\newblock \emph{arXiv preprint cs/0506075}.

\bibitem[{Qin et~al.(2024)Qin, Chen, Feng, Wu, Zhang, Li, Li, Che, and Yu}]{nlp1}
Libo Qin, Qiguang Chen, Xiachong Feng, Yang Wu, Yongheng Zhang, Yinghui Li, Min Li, Wanxiang Che, and Philip~S Yu. 2024.
\newblock \href {https://arxiv.org/pdf/2405.12819} {Large language models meet nlp: A survey}.
\newblock \emph{arXiv preprint arXiv:2405.12819}.

\bibitem[{Rajbhandari et~al.(2020)Rajbhandari, Rasley, Ruwase, and He}]{zero}
Samyam Rajbhandari, Jeff Rasley, Olatunji Ruwase, and Yuxiong He. 2020.
\newblock Zero: Memory optimizations toward training trillion parameter models.
\newblock In \emph{SC20: International Conference for High Performance Computing, Networking, Storage and Analysis}, pages 1--16. IEEE.

\bibitem[{Sener and Savarese(2018)}]{kgreedy}
Ozan Sener and Silvio Savarese. 2018.
\newblock \href {https://openreview.net/forum?id=H1aIuk-RW} {Active learning for convolutional neural networks: A core-set approach}.
\newblock In \emph{International Conference on Learning Representations}.

\bibitem[{Socher et~al.(2013)Socher, Perelygin, Wu, Chuang, Manning, Ng, and Potts}]{SST}
Richard Socher, Alex Perelygin, Jean Wu, Jason Chuang, Christopher~D Manning, Andrew~Y Ng, and Christopher Potts. 2013.
\newblock \href {https://aclanthology.org/D13-1170.pdf} {Recursive deep models for semantic compositionality over a sentiment treebank}.
\newblock In \emph{Proceedings of the 2013 conference on empirical methods in natural language processing}, pages 1631--1642.

\bibitem[{Team et~al.(2024{\natexlab{a}})Team, Mesnard, Hardin, Dadashi, Bhupatiraju, Pathak, Sifre, Rivi{\`e}re, Kale, Love et~al.}]{gemma}
Gemma Team, Thomas Mesnard, Cassidy Hardin, Robert Dadashi, Surya Bhupatiraju, Shreya Pathak, Laurent Sifre, Morgane Rivi{\`e}re, Mihir~Sanjay Kale, Juliette Love, et~al. 2024{\natexlab{a}}.
\newblock \href {https://arxiv.org/pdf/2403.08295} {Gemma: Open models based on gemini research and technology}.
\newblock \emph{arXiv preprint arXiv:2403.08295}.

\bibitem[{Team et~al.(2024{\natexlab{b}})Team, Riviere, Pathak, Sessa, Hardin, Bhupatiraju, Hussenot, Mesnard, Shahriari, Ram{\'e} et~al.}]{gemma2}
Gemma Team, Morgane Riviere, Shreya Pathak, Pier~Giuseppe Sessa, Cassidy Hardin, Surya Bhupatiraju, L{\'e}onard Hussenot, Thomas Mesnard, Bobak Shahriari, Alexandre Ram{\'e}, et~al. 2024{\natexlab{b}}.
\newblock \href {https://arxiv.org/pdf/2408.00118} {Gemma 2: Improving open language models at a practical size}.
\newblock \emph{arXiv preprint arXiv:2408.00118}.

\bibitem[{Tirumala et~al.(2024)Tirumala, Simig, Aghajanyan, and Morcos}]{D4}
Kushal Tirumala, Daniel Simig, Armen Aghajanyan, and Ari Morcos. 2024.
\newblock \href {https://proceedings.neurips.cc/paper_files/paper/2023/file/a8f8cbd7f7a5fb2c837e578c75e5b615-Paper-Datasets_and_Benchmarks.pdf} {D4: Improving llm pretraining via document de-duplication and diversification}.
\newblock \emph{Advances in Neural Information Processing Systems}, 36.

\bibitem[{Touvron et~al.(2023{\natexlab{a}})Touvron, Lavril, Izacard, Martinet, Lachaux, Lacroix, Rozi{\`e}re, Goyal, Hambro, Azhar et~al.}]{Llama}
Hugo Touvron, Thibaut Lavril, Gautier Izacard, Xavier Martinet, Marie-Anne Lachaux, Timoth{\'e}e Lacroix, Baptiste Rozi{\`e}re, Naman Goyal, Eric Hambro, Faisal Azhar, et~al. 2023{\natexlab{a}}.
\newblock \href {https://arxiv.org/pdf/2302.13971} {Llama: Open and efficient foundation language models}.
\newblock \emph{arXiv preprint arXiv:2302.13971}.

\bibitem[{Touvron et~al.(2023{\natexlab{b}})Touvron, Martin, Stone, Albert, Almahairi, Babaei, Bashlykov, Batra, Bhargava, Bhosale et~al.}]{llama2}
Hugo Touvron, Louis Martin, Kevin Stone, Peter Albert, Amjad Almahairi, Yasmine Babaei, Nikolay Bashlykov, Soumya Batra, Prajjwal Bhargava, Shruti Bhosale, et~al. 2023{\natexlab{b}}.
\newblock \href {https://arxiv.org/pdf/2307.09288} {Llama 2: Open foundation and fine-tuned chat models}.
\newblock \emph{arXiv preprint arXiv:2307.09288}.

\bibitem[{Yang et~al.(2024)Yang, Yang, Hui, Zheng, Yu, Zhou, Li, Li, Liu, Huang et~al.}]{qwen2}
An~Yang, Baosong Yang, Binyuan Hui, Bo~Zheng, Bowen Yu, Chang Zhou, Chengpeng Li, Chengyuan Li, Dayiheng Liu, Fei Huang, et~al. 2024.
\newblock \href {https://arxiv.org/pdf/2407.10671} {Qwen2 technical report}.
\newblock \emph{arXiv preprint arXiv:2407.10671}.

\bibitem[{Zeng et~al.(2023)Zeng, Yu, Gao, Meng, Goyal, and Chen}]{instructionf}
Zhiyuan Zeng, Jiatong Yu, Tianyu Gao, Yu~Meng, Tanya Goyal, and Danqi Chen. 2023.
\newblock \href {https://arxiv.org/pdf/2310.07641} {Evaluating large language models at evaluating instruction following}.
\newblock \emph{arXiv preprint arXiv:2310.07641}.

\bibitem[{Zhang et~al.(2023)Zhang, Dong, Li, Zhang, Sun, Wang, Li, Hu, Zhang, Wu et~al.}]{instruction}
Shengyu Zhang, Linfeng Dong, Xiaoya Li, Sen Zhang, Xiaofei Sun, Shuhe Wang, Jiwei Li, Runyi Hu, Tianwei Zhang, Fei Wu, et~al. 2023.
\newblock \href {https://arxiv.org/pdf/2308.10792} {Instruction tuning for large language models: A survey}.
\newblock \emph{arXiv preprint arXiv:2308.10792}.

\bibitem[{Zhang et~al.(2015)Zhang, Zhao, and LeCun}]{AGNEWS}
Xiang Zhang, Junbo Zhao, and Yann LeCun. 2015.
\newblock \href {https://proceedings.neurips.cc/paper/2015/file/250cf8b51c773f3f8dc8b4be867a9a02-Paper.pdf} {Character-level convolutional networks for text classification}.
\newblock \emph{Advances in neural information processing systems}, 28.

\bibitem[{Zhang et~al.(2024)Zhang, Wang, Ren, Li, Tiwari, Wang, and Qin}]{Pushing}
Yazhou Zhang, Mengyao Wang, Chenyu Ren, Qiuchi Li, Prayag Tiwari, Benyou Wang, and Jing Qin. 2024.
\newblock Pushing the limit of llm capacity for text classification.
\newblock \emph{arXiv preprint arXiv:2402.07470}.

\end{thebibliography}

\end{document}